\documentclass[10pt,twocolumn,letterpaper]{article}

\usepackage{cvpr}
\usepackage{times}
\usepackage{epsfig}
\usepackage{graphicx}
\usepackage{amsmath}
\usepackage{amssymb}
\usepackage{booktabs}
\usepackage[breaklinks=true,bookmarks=false]{hyperref}

 \cvprfinalcopy % *** Uncomment this line for the final submission

\def\cvprPaperID{720} % *** Enter the CVPR Paper ID here

\ifcvprfinal\fi
\begin{document}

%%%%%%%%% TITLE
\title{Hypercolumns for Object Segmentation and Fine-grained Localization}

\author{Bharath Hariharan\\
University of California\\
Berkeley\\
{\tt\small bharath2@eecs.berkeley.edu}
\and
Pablo Arbel\'{a}ez\\
Universidad de los Andes\\
Colombia\\
{\tt\small pa.arbelaez@uniandes.edu.co}
\and
Ross Girshick\\
Microsoft Research\\
Redmond\\
{\tt\small rbg@microsoft.com}
\and
Jitendra Malik\\
University of California\\
Berkeley\\
{\tt\small malik@eecs.berkeley.edu}
}

\maketitle
\thispagestyle{empty}

%%%%%%%%% ABSTRACT

\begin{abstract}
Recognition algorithms based on convolutional networks (CNNs) typically use the output of the last layer as a feature representation. However, the information in this layer may be too coarse spatially to allow precise localization. On the contrary, earlier layers may be precise in localization but will not capture semantics. To get the best of both worlds, we define the \emph{hypercolumn} at a pixel as the vector of activations of all CNN units above that pixel. Using hypercolumns as pixel descriptors, we show results on three fine-grained localization tasks: simultaneous detection and segmentation~\cite{BharathECCV2014}, where we improve state-of-the-art from \emph{49.7} mean AP$^r$~\cite{BharathECCV2014} to \emph{60.0}, keypoint localization, where we get a 3.3 point boost over~\cite{GkioxariArxiv2014}, and part labeling, where we show a 6.6 point gain over a strong baseline.  
\end{abstract}

%%%%%%%%% BODY TEXT
\section{Introduction}
Features based on convolutional networks (CNNs)~\cite{LecunNC1989} have now led to the best results on a range of vision tasks: image classification~\cite{KrizhevskyNIPS2012,SimonyanArxiv2014a}, object segmentation and detection~\cite{GirshickCVPR2014,BharathECCV2014}, action classification~\cite{SimonyanArxiv2014b}, pose estimation~\cite{TompsonNIPS2014} and fine-grained category recognition~\cite{ZhangECCV2014,BransonBMVC2014}. We have thus moved from the era of HOG and SIFT to the era of convolutional network features. Therefore, understanding these features and how best to exploit them is of wide applicability.

Typically, recognition algorithms use the output of the last layer of the CNN. This makes sense when the task is assigning category labels to images or bounding boxes: the last layer is the most sensitive to category-level semantic information and the most invariant to ``nuisance" variables such as pose, illumination, articulation, precise location and so on. However, when the task we are interested in is finer-grained, such as one of segmenting the detected object or estimating its pose, these nuisance variables are precisely what we are interested in. For such applications, the top layer is thus \emph{not} the optimal representation.

%This makes sense for category-level recognition: owing to the sequence of non-linearities and pooling operations, this layer is the most sensitive to category-level semantic information and the most invariant to changes in pose, illumination, articulation, precise location and so on. However, this invariance to nuisance variables also means that if these nuisance variables are what we are interested in, this layer is \emph{not} the optimal representation. For instance, if we are interested in segmentation or pose estimation, we need pixel-precise localization. However, the top layer units have very large, image-sized receptive fields and hence probably contain only coarse localization information.

The information that is generalized over in the top layer is present in intermediate layers, but intermediate layers are also much less sensitive to semantics. For instance, bar detectors in early layers might localize bars precisely, but cannot discriminate between bars that are horse legs and bars that are tree trunks. This observation suggests that reasoning at multiple levels of abstraction and scale is necessary, mirroring other problems in computer vision where reasoning across multiple levels has proven beneficial. For example, in optical flow, coarse levels of the image pyramid are good for correspondence, but finer levels are needed for accurate measurement, and a multiscale strategy is used to get the best of both worlds~\cite{BroxECCV2004}. 

\begin{figure}
\includegraphics[width=0.45\textwidth]{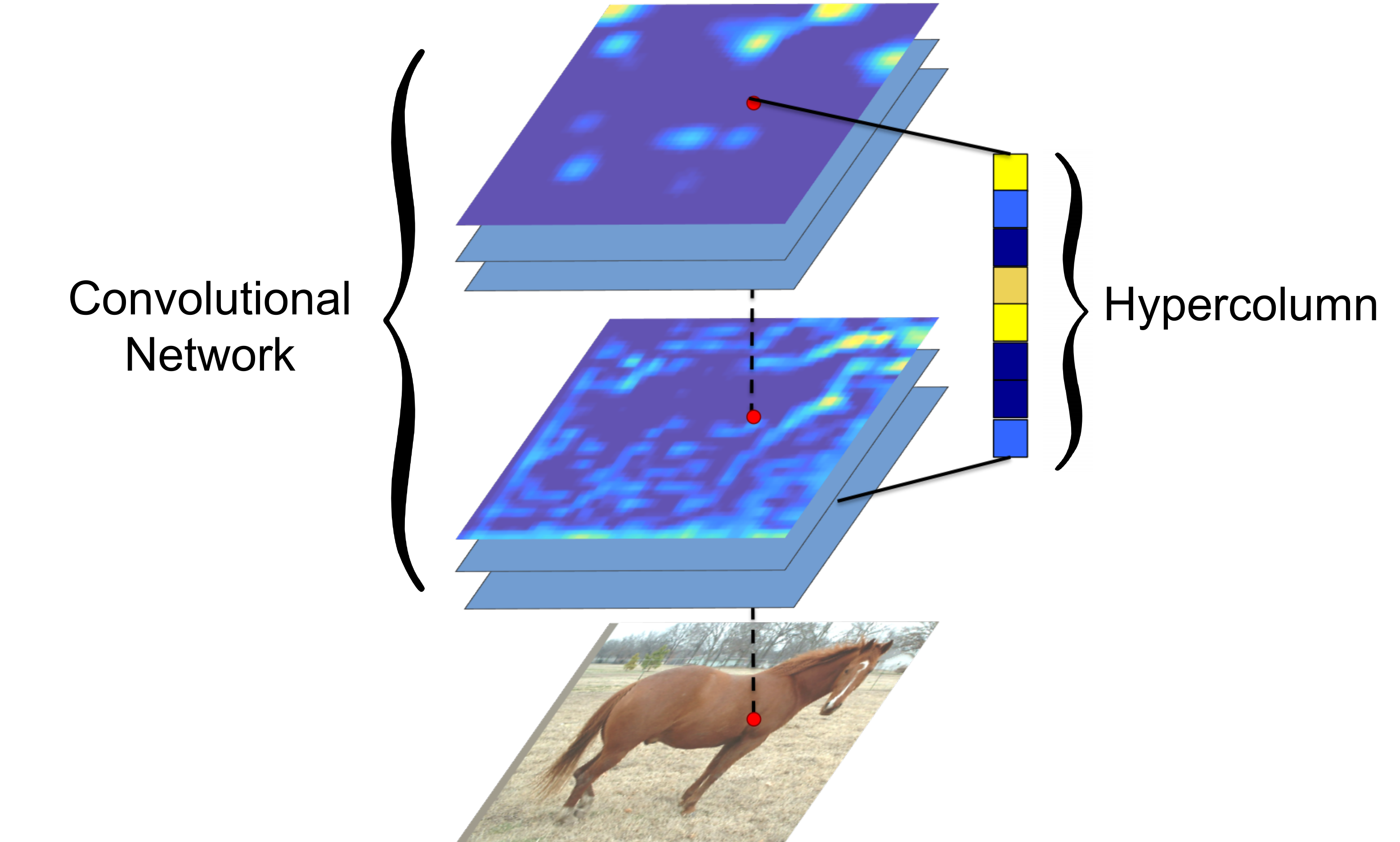}
\caption{The hypercolumn representation. The bottom image is the input, and above it are the  feature maps of different layers in the CNN. The hypercolumn at a pixel is the vector of activations of all units that lie above that pixel.}\label{fig:hyp}
\end{figure}

In this paper, we think of the layers of a convolutional network as a non-linear counterpart of the image pyramids used in optical flow and other vision tasks. Our hypothesis is that the information of interest is distributed over \emph{all} levels of the CNN and should be exploited in this way. We define the ``hypercolumn" at a given input location as the outputs of all units above that location at all layers of the CNN, stacked into one vector. (Because adjacent layers are strongly correlated, in practice we need not consider all layers but can simply sample a few.) Figure~\ref{fig:hyp} shows a visualization of the idea. We borrow the term ``hypercolumn" from neuroscience, where it is used to describe a set of V1 neurons sensitive to edges at multiple orientations and multiple frequencies arranged in a columnar structure~\cite{HubelTJoP1962}. 
However, our hypercolumn includes not just edge detectors but also more semantic units and is thus a more general notion.

We show the utility of the hypercolumn representation on two kinds of problems that require precise localization. The first problem is simultaneous detection and segmentation (SDS)~\cite{BharathECCV2014}, where the aim is to both detect and segment every instance of an object category in the image. The second problem deals with detecting an object and localizing its parts. We consider two variants of this: one, locating the keypoints~\cite{YangTPAMI2013}, and two, segmenting out each part~\cite{YamaguchiCVPR2012, YamaguchiICCV2013, BoCVPR2011, LuoICCV2013}.

We present a general framework for tackling these and other fine-grained localization tasks by framing them as pixel classification and using hypercolumns as pixel descriptors. We formulate our entire system as a neural network, allowing end-to-end training for particular tasks simply by changing the target labels. Our empirical results are:
\begin{enumerate}
\item On SDS, the previous state-of-the-art was 49.7 mean AP$^r$~\cite{BharathECCV2014}. Substituting hypercolumns into the pipeline of~\cite{BharathECCV2014} improves this to \textbf{52.8}. We also propose a more efficient pipeline that allows us to use a larger network, pushing up the performance to \textbf{60.0}. 
\item On keypoint prediction, we show that a simple keypoint prediction scheme using hypercolumns achieves a \textbf{3.3} point gain in the APK metric~\cite{YangTPAMI2013} over prior approaches working with only the top layer features~\cite{GkioxariArxiv2014}. While there isn't much prior work on labeling parts of objects, we show that the hypercolumn framework is significantly better (by \textbf{6.6} points on average) than a strong baseline based on the top layer features.
\end{enumerate}

%As discussed above, compared to the top-most layers of the network, we expect the hypercolumn representation to be much better at fine-grained localization tasks. We consider three such tasks as our testbed: 
%
%\begin{enumerate}
%\item As our primary application we show results on simultaneous detection and segmentation (SDS)~\cite{BharathECCV2014}, where the task is to detect and segment every instance of a category in an image. We use hypercolumns to predict figure-ground masks for detected objects. We show a \httilde3 point gain on AP$^r$ (\httilde8 point gain using a stricter overlap threshold) compared to~\cite{BharathECCV2014}. Using a more sophisticated network increases the gain to \httilde6 points.
%\item We show that a relatively simple keypoint prediction system using hypercolumns achieves a 1.8 point gain in the APK metric~\cite{YangTPAMI2013} over~\cite{GkioxariArxiv2014}.
%\item Finally, we demonstrate the ability to segment out not just the object but also parts of the object.  In this challenging task, we show large gains of \httilde6 percentage points over a strong baseline.
%\end{enumerate}
\section{Related work}
\noindent\textbf{Combining features across multiple levels: }
Burt and Adelson introduced Laplacian pyramids~\cite{BurtTOC1983}, a representation that is widely used in computer vision. Koenderink and van Doorn~\cite{KoenderinkBC1987} used ``jets", which are sets of partial derivatives of intensity up to a particular order, to estimate edge orientation, curvature, etc. Malik and Perona~\cite{MalikJOSA1990} used the output of a bank of filters as a representation for texture discrimination. This representation also proved useful for optical flow~\cite{WeberIJCV1995} and stereo~\cite{JonesECCV1992}. While the filter banks in these works cover multiple scales, they are still restricted to simple linear filters, whereas many of the features in the hypercolumn representation are highly non-linear functions of the image.

There has also been work in convolutional networks that combines multiple levels of abstraction and scale. Farabet et al.~\cite{FarabetTPAMI2013} combine CNN outputs from multiple scales of an image to do semantic segmentation. Tompson et al.~\cite{TompsonNIPS2014} use a similar idea for detecting parts and estimating pose. However, the features being combined still come from the same level of the CNN and hence have similar invariance. Sermanet et al.~\cite{SermanetCVPR2013} combine subsampled intermediate layers with the top layer for pedestrian detection. In contrast, since we aim for precise localization, we maintain the high resolution of the lower layers and upsample the higher layers instead. In contemporary work, Long et al.~\cite{LongCVPR2015} also use multiple layers for their fully convolutional semantic segmentation system. 

~\\\noindent\textbf{Detection and segmentation: } The task of simultaneous detection and segmentation task, introduced in~\cite{BharathECCV2014}, requires one to detect and segment every instance of a category in the image. SDS differs from classical bounding box detection in its requirement of a segmentation and from classical semantic segmentation in its requirement of separate instances. There has been other prior work on segmenting out instances of a category, mostly starting from bounding box detections. Borenstein and Ullman~\cite{BorensteinECCV2002} first suggested the idea of using class-specific knowledge for segmentation. Yang et al.~\cite{YangTPAMI2012} use figure ground masks associated with DPM detectors~\cite{FelzenszwalbPAMI2010} to segment out detected objects and reason about depth orderings. Parkhi et al.~\cite{ParkhiICCV2011} use color models extracted from the detected cat and dog heads to segment them out. Dai and Hoiem~\cite{DaiCVPR2012} generalize this reasoning to all categories. Fidler et al.~\cite{FidlerCVPR2013} and Dong et al.~\cite{DongECCV2014} combine object detections from DPM~\cite{FelzenszwalbPAMI2010} with semantic segmentation outputs from O$_2$P~\cite{CarreiraECCV2012} to improve both systems. Current leading methods use CNNs to score bottom-up object proposals, both for object detection~\cite{GirshickCVPR2014} and for SDS~\cite{BharathECCV2014,DaiCVPR2015}. 

~\\\noindent\textbf{Pose estimation and part labeling: } Current best performers for pose estimation are based on CNNs. Toshev and Szegedy~\cite{ToshevCVPR2014} use a CNN to regress to keypoint locations. Tompson et al.~\cite{TompsonNIPS2014} show large improvements over state-of-the-art by predicting a heatmap for each keypoint, where the value of the heatmap at a location is the probability of the keypoint at that location. These algorithms show results in the setting where the rough location of the person is known. Yang and Ramanan~\cite{YangTPAMI2013} propose a more  realistic setting where the location of the person is not known and one has to both detect the person and identify his/her keypoints. Gkioxari et al.~\cite{GeorgiaBharathCVPR2014b} show some results in this setting using HOG-based detectors, but in their later work~\cite{GkioxariArxiv2014} show large gains using CNNs.

Related to pose estimation is the task of segmenting out the different parts of a person, a task typically called ``object parsing". Yamaguchi et al.~\cite{YamaguchiCVPR2012, YamaguchiICCV2013} parse fashion photographs into clothing items. There has also been work on parsing pedestrians~\cite{BoCVPR2011, LuoICCV2013}. Ionescu et al.~\cite{IonescuCVPR2014} jointly infer part segmentations and pose. However, the setting is typically tightly cropped bounding boxes of pedestrians, while we are interested in the completely unconstrained case. 

\section{Pixel classification using hypercolumns}\label{sec:hypercolumn}
\noindent\textbf{Problem setting:}
We assume an object detection system that gives us a set of detections. Each detection comes with a bounding box, a category label and a score (and sometimes an initial segmentation hypothesis). The detections have already been subjected to non-maximum suppression. For every detection, we want to segment out the object, segment its parts or predict its keypoints.

For each task, we expand the bounding box of the detection slightly and predict a heatmap on this expanded box. The type of information encoded by this heatmap depends on the particular task. For segmentation, the heatmap encodes the probability that a particular location is inside the object. For part labeling, we predict a separate heatmap for each part, where each heatmap is the probability a location belongs to that part. For keypoint prediction, again we output a separate heatmap for each keypoint, with each heatmap encoding the probability that the keypoint is at a particular location.

In each case, we predict a $50 \times 50$ heatmap that we resize to the size of the expanded bounding box and splat onto the image. Thus, in our framework, these diverse fine-grained localization problems are addressed as the unified task of assigning a probability to each of the $50 \times 50$ locations or, in other words, of classifying each location. We solve this classification problem using the hypercolumn representation as described in detail below.
\vspace{2mm}
~\\\noindent\textbf{Computing the hypercolumn representation:}
We take the cropped bounding box, resize it to a fixed size and feed it into a CNN as in~\cite{GirshickCVPR2014}. For each location, we extract features from a set of layers by taking the outputs of the units that are ``above" the location (as shown in Figure~\ref{fig:hyp}). All the intermediate outputs in a CNN are feature maps (the output of a fully connected layer can be seen as a $1\times 1$ feature map). However, because of subsampling and pooling operations in the CNN, these feature maps need not be at the same resolution as the input or the target output size. So which unit lies above a particular location is ambiguous. We get around this by simply resizing each feature map to the size we want with bilinear interpolation. If we denote the feature map by $\mathbf{F}$ and the upsampled feature map by $\mathbf{f}$, then the feature vector for the $i$th location has the form:
\begin{eqnarray}
\mathbf{f}_i  = \sum_k \alpha_{ik} \mathbf{F}_{k}\label{eqn1}
\end{eqnarray}
$\alpha_{ik}$ depends on the position of $i$ and $k$ in the box and feature map respectively. %In practice, since our output and input sizes are fixed, the $\alpha$'s, which depend only on the size of the feature map and the target size, can be precomputed. 

We concatenate features from some or all of the feature maps in the network into one long vector for every location which we call the hypercolumn at that location. As an example, using \emph{pool2} (256 channels), \emph{conv4} (384 channels) and \emph{fc7} (4096 channels) from the architecture of~\cite{KrizhevskyNIPS2012} would lead to a 4736 dimensional vector. 
\vspace{2mm}
~\\\noindent\textbf{Interpolating into a grid of classifiers:}
Because these feature maps are the result of convolutions and poolings, they do not encode any information about where in the bounding box a given pixel lies. However, location can be an important feature. For instance, in a person bounding box, the head is more likely to be at the top of the bounding box than at the bottom. Thus a pixel that looks like a nose should be considered as part of the person if it occurs at the top of the box and should be classified as background otherwise. The reasoning should be the opposite for a foot-like pixel. This is a highly non-linear effect of location, and such reasoning cannot be achieved simply by a location-specific bias. (Indeed, our classifiers include $(x,y)$ as features but assign negligible weight to them). Such reasoning requires different classifiers for each location.

Location is also needed to make better use of the features from the fully connected layers at the top. Since these features are shared by all the locations in the bounding box, they can at best contribute a global instance-specific bias. However, with a different classifier at each location, we can have a separate instance-specific bias for each location. Thus location-specific classifiers in conjunction with the global, instance-level features from the fully connected layer produce an instance-specific prior.

The simplest way to get a location-specific classifier is to train separate classifiers for each of the $50 \times 50$ locations. However, doing so has three problems. One, it dramatically reduces the amount of data each classifier sees during training. In our training sets, some categories may have only a few hundred instances, while the dimensionality of the feature vector is of the order of several thousand. Thus, having fewer parameters and more sharing of data is necessary to prevent overfitting. Two, training this many classifiers is computationally expensive, since we will have to train 2500 classifiers for 20 categories. Three, while we do want the classifier to vary with location, the classifier should change slowly: two adjacent pixels that are similar to each other in appearance should also be classified similarly.

Our solution is to train a coarse $K \times K$ grid of classifiers and interpolate between them.  In our experiments we use $K=5$ or 10. For the interpolation, we use an extension of bilinear interpolation where we interpolate a grid of \emph{functions} instead of a grid of \emph{values}. Concretely, each classifier in the grid is a function $g_k(\cdot)$ that takes in a feature vector and outputs a probability between 0 and 1. We use this coarse grid of functions to define the function $h_i$ at each pixel $i$ as a linear combination of the nearby grid functions, analogous to Equation~\ref{eqn1}:
%, We use this coarse grid of functions to define a function at every pixel. Denote the function at pixel $i$ as $h_i(\cdot)$. Then $h_i(\cdot)$ is obtained by bilinearly interpolating the grid functions $g_k(\cdot)$, analogous to equation~\ref{eqn1}: 
%To simplify exposition, let us consider the one-dimensional case first. In this case we have the grid classifiers defined at discrete points in an interval (say $[0,1]$), and we want to define a function at every point in $[0, 1]$. Consider a point $p$, and denote the function at $p$ as $h_p$. Let the two adjacent grid locations be $q_1$ and $q_2$, with $q_1<q_2$, and let the corresponding grid functions be $g_1$ and $g_2$. Then we define $h_p$ as:
%\begin{equation}
%h_p(\cdot) = \frac{q_2 - p}{q_2 - q_1}g_1(\cdot)+\frac{p-q_1}{q_2 - q_1}g_2(\cdot)
%\end{equation}
%
%In other words, $h_p$ is simply a linear combination of the grid functions, with the coefficients being non-zero only for the adjacent grid functions. The equation in two dimensions is similar, and in general we can simply write, for a pixel $i$ :
 \begin{equation}
h_i(\cdot) = \sum_k \alpha_{ik} g_k (\cdot)
\end{equation}

If the feature vector at the $i$th pixel is $\mathbf{f}_i$, then the score of the $i$th pixel is:
\begin{eqnarray}
p_i = \sum_k \alpha_{ik} g_k(\mathbf{f}_i) =\sum_k\alpha_{ik} p_{ik}\label{eqn:heatmapinterp}
\end{eqnarray}
where $p_{ik}$ is the probability output by the $k$th classifier for the $i$th pixel. Thus, at test time we run all our $K^2$ classifiers on all the pixels. Then, at each pixel, we linearly combine the outputs of all classifiers at that pixel using the above equation to produce the final prediction. Note that the coefficients of the linear combination depend on the location.  

Training this interpolated classifier is a hard optimization problem. We use a simple heuristic and ignore the interpolation at train time, using it only at test time.We divide each training bounding box into a $K \times K$ grid. The training data for the $k$th classifier consists only of pixels from the $k$th grid cell across all training instances. Each classifier is trained using logistic regression. This training methodology does not directly optimize the loss we would encounter at test time, but allows us to use off-the-shelf code such as liblinear~\cite{FanJMLR2008} to train the logistic regressor.

~\\\noindent\textbf{Efficient classification using convolutions and upsampling: }
Our system requires us to resize every feature map to $50 \times 50$ and then classify each location. But resizing feature maps with hundreds of channels can be expensive. However, we know we are going to run several linear classifiers on top of the hypercolumn features and we can use this knowledge to save computation as follows: each feature map with $c$ channels will give rise to a $c$-dimensional block of features in the hypercolumn representation of a location, and this block will have a corresponding block of weights in the classifiers. Thus if $\mathbf{f}_i$ is the feature vector at location $i$, then $\mathbf{f}_i$ will be composed of blocks $\mathbf{f}^{(j)}_i$ corresponding to the $j$th feature map. A linear classifier $\mathbf{w}$ will decompose similarly. The dot product between $\mathbf{w}$ and $\mathbf{f}_i$ can then be written as:
\begin{eqnarray}
\mathbf{w}^T\mathbf{f}_i = \sum_j \mathbf{w}^{(j)T} \mathbf{f}^{(j)}_i
\end{eqnarray}

The $j$th term in the decomposition corresponds to a linear classifier on top of the upsampled $j$th feature map. However, since the upsampling is a linear operation, we can first apply the classifier and then upsample using Equation~\ref{eqn1}:
\begin{eqnarray}
\mathbf{f}^{(j)}_i  = \sum_k \alpha^{(j)}_{ik} \mathbf{F}^{(j)}_{k} \\
\mathbf{w}^{(j)T}\mathbf{f}^{(j)}_i = \sum_k \alpha^{(j)}_{ik} \mathbf{w}^{(j)T}\mathbf{F}^{(j)}_{k}
\end{eqnarray}
We note that this insight was also used by Barron et al.~\cite{BarronICCV2013} in their volumetric semantic segmentation system.

Observe that applying a classifier to each location in a feature map is the same as a $1 \times 1$ convolution. Thus, to  run a linear classifier on top of hypercolumn features, we break it into blocks corresponding to each feature map, run $1\times 1$ convolutions on each feature map to produce score maps, upsample all score maps to the target resolution, and sum.

We consider a further modification to this pipeline where we replace the $1 \times 1$ convolution with a general $n \times n$ convolution. This corresponds to looking not only at the unit directly above a pixel but also the neighborhood of the unit. This captures the pattern of activations of a whole neighborhood, which can be more informative than a single unit, especially in the lower layers of the network.%can be useful, especially in the lower layers of the network, where the pattern of activations of a neighborhood of units is more informative than the activation of a single unit.

\begin{figure}
\includegraphics[width=0.5\textwidth]{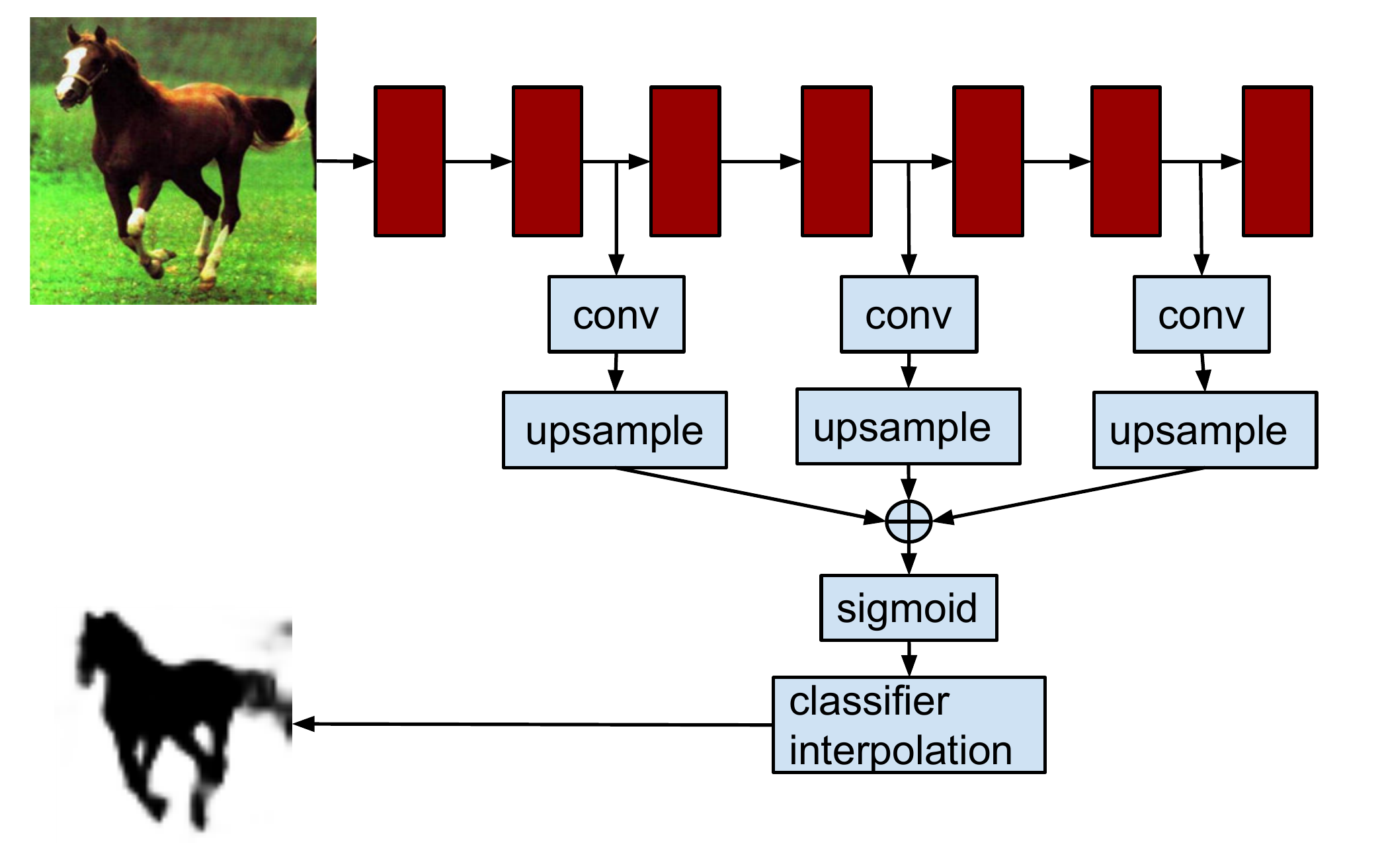}
\caption{Representing our hypercolumn classifiers as a neural network. Layers of the original classification CNN are shown in red, and layers that we add are in blue.}\label{fig:hypercolumnnet}
\end{figure}

 ~\\\noindent\textbf{Representation as a neural network: }
We can write our final hypercolumn classifier using additional layers grafted onto the original CNN as shown in Figure~\ref{fig:hypercolumnnet}. For each feature map, we stack on an additional convolutional layer. Each such convolutional layer has $K^2$ channels, corresponding to the $K^2$ classifiers we want to train. We can choose any kernel size for the convolutions as described above, although for fully connected layers that produce $1 \times 1$ feature maps, we are restricted to $1 \times 1$ convolutions. We take the outputs of all these layers, upsample them using bilinear interpolation and sum them. Finally, we pass these outputs through a sigmoid, and combine the $K^2$ heatmaps using  equation~\ref{eqn:heatmapinterp} to give our final output. Each operation is differentiable and can be back-propagated over. 

Representing our pipeline as a neural network allows us to train the whole network (including the CNN from which we extract features) for this task. For such training, we feed in the target $50 \times 50$ heatmap as a label. The loss is the sum of logistic losses (or equivalently, the sum of the negative log likelihoods) over all the $50 \times 50$ locations. We found that treating the sigmoids, the linear combination and the log likelihood as a single composite function and computing the gradient with respect to that led to simpler, more numerically stable expressions.
%Because computing the gradient of the negative log likelihood and then backpropagating it through a sigmoid is prone to numerical instabilities, we treat the sigmoids, the linear combination, and the negative log likelihood as a single composite function, and directly compute the gradient with respect to that. This leads to more numerically stable, simplified expressions. 
Instead of training the network from scratch, we use a pretrained network and finetune, i.e., do backpropagation with a small learning rate. Finally, this representation as a neural network also allows us to train the grid classifiers together and use classifier interpolation during training, instead of training separate grid classifiers independent of each other.

%~\\\noindent\textbf{Using the correct bounding box: }
%Our system predicts a heatmap on the bounding box, and so it cannot recover if the bounding box is inaccurate. A wrong bounding box affects us in two ways: if the box is too small, then our heatmap will not include object pixels that lie outside the box. The box also sets the coordinate system for our grid classifiers and our features. We therefore use the bounding box regression proposed by Girshick et al.~\cite{GirshickCVPR2014} to predict a better bounding box, and then make our predictions on this new bounding box. 
%
 
 ~\\\noindent\textbf{Training classifiers for segmentation and part localization:}
For each category we take bottom-up MCG candidates~\cite{ArbelaezCVPR2014} that overlap a ground truth instance by 70\% or more. For each such candidate, we find the ground truth instance it overlaps most with, and crop that ground truth instance to the expanded bounding box of the candidate. Depending on the task we are interested in (SDS, keypoint prediction or part labeling), we then use the labeling of the cropped ground truth instance to label locations in the expanded bounding box as positive or negative. For SDS, locations inside the instance are considered positive, while locations outside are considered negative. For part labeling, locations inside a \emph{part} are positive and all other locations are negative. For keypoint prediction, the true keypoint location is positive and locations outside a certain radius (we use 10\% of the bounding box diagonal) of the true location are labeled negative.  

%\begin{eqnarray}
%loss(\mathbf{s}) &= -\sum_i y_i \log \sum_k \alpha_{ik} p_{ik} \nonumber \\
%		&-\sum_i (1-y_i) \log (1-\sum_k \alpha_{ik} p_{ik})\\
%&=-\sum_i y_i \log \sum_k \alpha_{ik} p_{ik} \nonumber \\
%&-\sum_i(1-y_i) \log \sum_k \alpha_{ik} (1-p_{ik}) \\
%\end{eqnarray}
%Here $p_{ik} = \sigma(s_{ik}) = 1/(1+e^{-s_{ik}})$ and we have used the fact that the interpolation weights $\alpha_{ik}$ sum to 1 for fixed $i$. Taking the derivative with respect to $s_{ik}$ gives us:
%\begin{eqnarray}
%\frac{\partial loss}{\partial s_{ik}} &= -\sum_i y_i \frac{\alpha_{ik}}{\sum_k \alpha_{ik} p_{ik} } \frac{\partial p_{ik}}{\partial s_{ik}} \nonumber \\
%			&+sum_i (1-y_i) \frac{\alpha_{ik}}{\sum_k \alpha_{ik}(1-p_{ik})} \frac{\partial p_{ik}}{\partial s_{ik}}\\
%\frac{\partial loss}{\partial s_{ik}} &=  -\sum_i y_i \frac{\alpha_{ik} p_{ik} (1-p_{ik})}{p_i} + \sum_i (1-y_i) \frac{\alpha_{ik} p_{ik} (1-p_{ik})}{1 - p_i}			
%\end{eqnarray} 

\section{Experiments on SDS}\label{sec:SDSintro}
Our first testbed is the SDS task. Our baseline for this task is the algorithm presented in~\cite{BharathECCV2014}. This pipeline scores bottom-up region proposals from~\cite{ArbelaezCVPR2014} using CNN features computed on both the cropped bounding box of the region and the cropped region foreground. The regions are subjected to non-max suppression. Finally, the surviving candidates are refined using figure-ground predictions based on the top layer features.

As our first system for SDS, we use the same pipeline as above, but replace the refinement step with one based on hypercolumns. (We also add a bounding box regression step~\cite{GirshickCVPR2014} so as to start from the best available bounding box). We present results with this pipeline in section~\ref{sec:SDS1}, where we show that hypercolumn-based refinement is significantly better than the refinement in~\cite{BharathECCV2014}, and is especially accurate when it comes to capturing fine details of the segmentation. We also evaluate several ablations of our system to unpack this performance gain. For ease of reference, we call this \textbf{System 1}.

\begin{figure}
\centering
\includegraphics[width=0.5\textwidth]{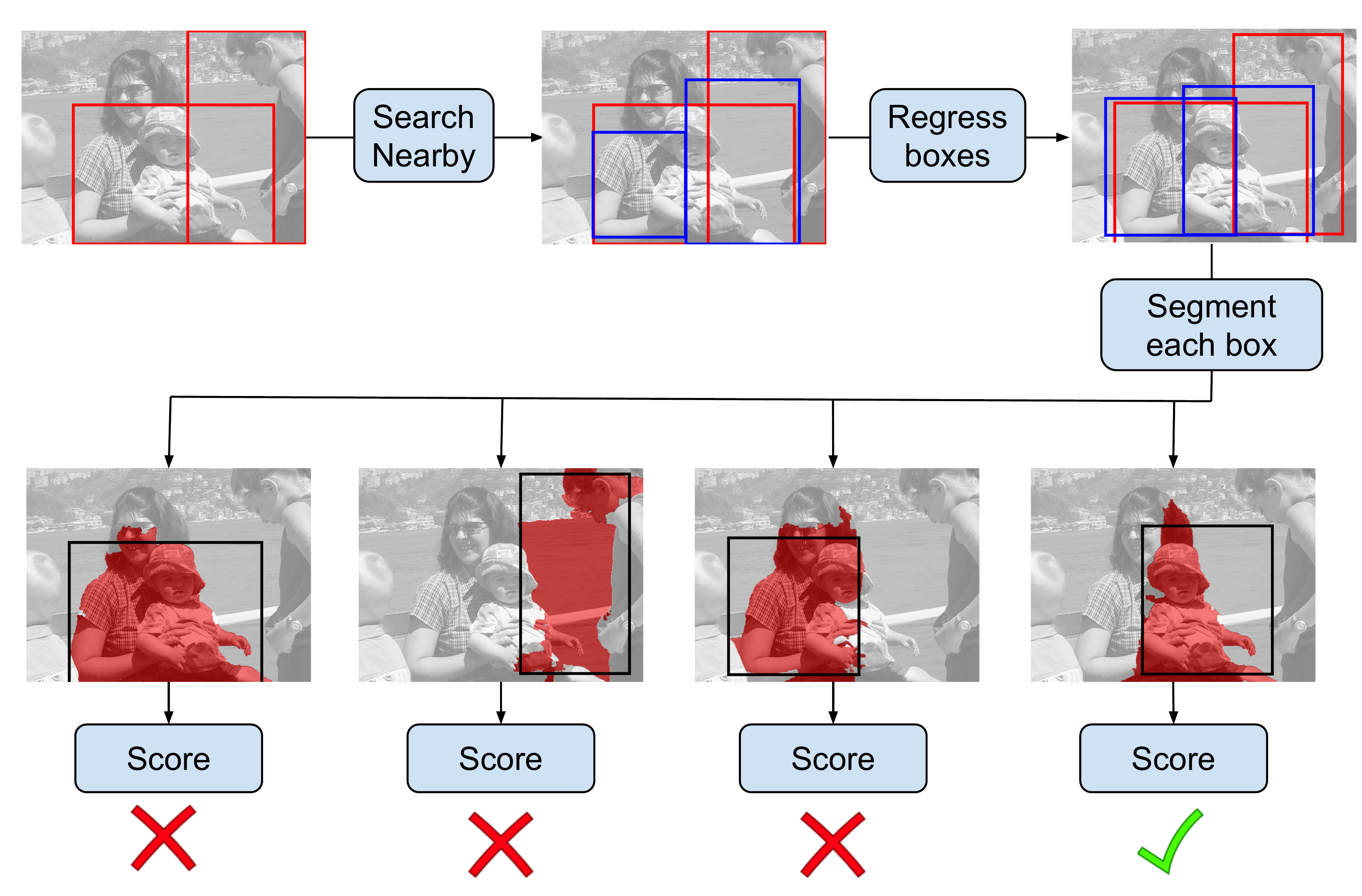}
\caption{An alternative pipeline for SDS starting from bounding box detections (Section~\ref{sec:SDSintro})}\label{fig:rescore}
\end{figure}

One issue with this system is its computational cost. Extracting features from region foregrounds is expensive and doubles the time taken. Further, while CNN-based bounding box detection~\cite{GirshickCVPR2014} can be speeded up dramatically using approaches such as~\cite{HeECCV2014}, no such speedups exist for region classification. To address these drawbacks, we propose as our second system the pipeline shown in Figure~\ref{fig:rescore}. This pipeline starts with bounding box detections after non-maximum suppression. We expand this set of detections by adding nearby high-scoring boxes that were removed by non-maximum suppression but may be better localized (explained in detail below). This expanded set is only twice as large as the original set, and about two orders of magnitude smaller than the full set of bottom-up proposals. For each candidate in this set, we predict a segmentation, and score this candidate using CNN features computed on the segmentation. Because region-based features are computed only on a small set, the pipeline is much more efficient.  We call this system \textbf{System 2}.

This pipeline relies crucially on our ability to predict a good segmentation from just bounding boxes. We use hypercolumns to make this prediction. In section~\ref{sec:SDS2}, we show that these predictions are accurate, and significantly better than predictions based on the top layer of the CNN. 

Finally, the efficiency of this pipeline also allows us to experiment with larger but more expressive architectures. While ~\cite{BharathECCV2014} used the architecture proposed by Krizhevsky et al.~\cite{KrizhevskyNIPS2012} (referred to as ``T-Net" henceforth, following~\cite{GirshickCVPR2014Arxiv}) for both the box features and the region features, we show in section~\ref{sec:SDS2} that the architecture proposed by Simonyan and Zisserman~\cite{SimonyanArxiv2014a} (referred to as ``O-Net" henceforth~\cite{GirshickCVPR2014Arxiv}) is significantly better.

\subsection{System 1: Refinement using hypercolumns}\label{sec:SDS1}
In our first set of experiments, we compare a hypercolumn-based refinement to that proposed in~\cite{BharathECCV2014}. We use the ranked hypotheses produced by~\cite{BharathECCV2014} and refine each hypothesis using hypercolumns. For the CNN, we use the same network that was used for the region classification (described as \textbf{C} in~\cite{BharathECCV2014}). This network consists of two pathways, each based on T-Net. It takes in both the cropped bounding box as well as the cropped foreground. For the hypercolumn representation we use the top-level \emph{fc7} features, the \emph{conv4} features from both pathways using a $1\times1$ neighborhood, and the \emph{pool2} features from the box pathway with a $3 \times 3$ neighborhood. We choose these layers because they are spread out evenly in the network and capture a diverse set of features. In addition, for each location, we add as features a 0 or 1 encoding if the location was inside the original region candidate, and a coarse $10\times 10$ discretization of the original candidate flattened into a 100-dimensional vector. This is to be commensurate with~\cite{BharathECCV2014} where these features were used in the refinement step. We use a $10 \times 10$ grid of classifiers. As a last step, we project our predictions to superpixels by averaging the prediction over each superpixel. We train on VOC2012 Train and evaluate on VOC2012 Val.

Table~\ref{table:hypercolumn-basic} shows the results of our experiments. The first two columns show the performance reported in~\cite{BharathECCV2014} with and without the refinement step. ``Hyp" is the result we get using hypercolumns, without bounding box regression or finetuning. Our mean AP$^r$ at 0.5 is \textbf{1.5} points higher, and at 0.7 is \textbf{6.3} points higher, indicating that our refinement is much better than that of~\cite{BharathECCV2014} and is a large improvement over the original candidate. Bounding box regression and finetuning the network both provide significant gains, and with both of these, our mean AP$^r$ at 0.5 is \textbf{3.1} points higher and at 0.7 is \textbf{8.4} points higher than~\cite{BharathECCV2014}. 

Table~\ref{table:hypercolumn-basic} also shows the results of several ablations of our model (all without bounding box regression or finetuning):
\begin{enumerate}
\item \emph{Only fc7} uses only \emph{fc7} features and is thus similar to the refinement step in~\cite{BharathECCV2014}. We include this baseline to confirm that we can replicate those results.
\item \emph{fc7+pool2, fc7+conv4} and \emph{pool2+conv4} are refinement systems that use hypercolumns but leave out features from \emph{conv4}, \emph{pool2} and \emph{fc7} respectively. Each of these baselines performs worse than our full system. In each case the difference is statistically significant at a confidence threshold of $0.05$, computed using paired sample permutation tests.
\item The \emph{$1 \times 1$, $2 \times 2$} and \emph{$5 \times 5$} models use different grid resolutions, with the $1 \times 1$ grid amounting to a single classifier. There is a significant loss in performance (2.4 points at 0.7 overlap) when using a $1\times1$ grid. However this baseline still outperforms~\cite{BharathECCV2014} indicating that even without our grid classifiers (and without \emph{fc7}, since the global \emph{fc7} features are ineffectual without the grid), the hypercolumn representation by itself is quite powerful. A $5\times5$ grid is enough to recover full performance.
\end{enumerate}

Finally, following~\cite{BharathECCV2014}, we take our Hyp+FT+bbox-reg system and use the pasting scheme of~\cite{CarreiraECCV2012} to obtain a semantic segmentation. We get a mean IU of \textbf{54.6} on VOC2012 Segmentation Test, 3 points higher than~\cite{BharathECCV2014} (51.6 mean IU).

\subsection{System 2: SDS from bounding box detections}\label{sec:SDS2}
For our experiments with System 2, we use the detections of R-CNN~\cite{GirshickCVPR2014} as the starting point.  R-CNN uses CNNs to classify bounding box proposals from selective search. We use the final output after non-max suppression and bounding box regression. However, to allow direct comparison with our previous experiments, we retrained R-CNN to work with box proposals from MCG~\cite{ArbelaezCVPR2014}. We do all training on VOC2012 Train.

We first evaluate our segmentation predictions. As before, we use the same network as the detector to compute the hypercolumn transform features. We first experiment with the T-Net architecture. We use the layers \emph{fc7}, \emph{conv4} with a neighborhood of 1, and \emph{pool2} with a neighborhood of 3. For computational reasons we do not do any finetuning. We use superpixel projection as before.

We show results in Table~\ref{table:bbox}. Since we use only one network operating on bounding boxes instead of two working on both the box and the region, we expect a drop in performance. We find that this is the case, but the loss is small: we get a mean AP$^r$ of 49.1 at 0.5 and 29.1 at 0.7, compared to 51.9 and 32.4 when we have the region features. In fact, our performance is nearly as good as~\cite{BharathECCV2014} at 0.5 and about 4 points better at 0.7, and we get this accuracy starting from just the bounding box. 

\begin{figure}
\centering
\includegraphics[height=4\baselineskip]{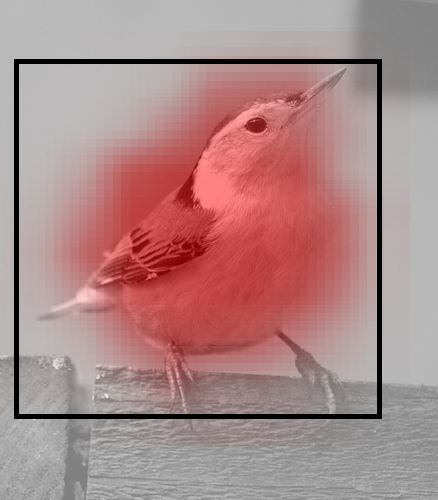}
\includegraphics[height=4\baselineskip]{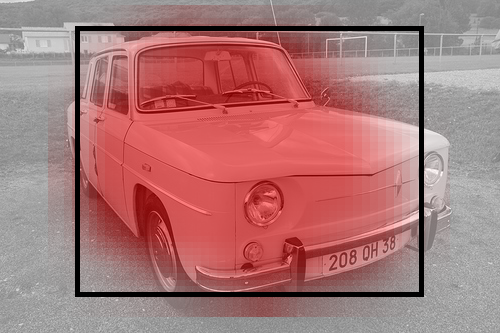}
\includegraphics[height=4\baselineskip]{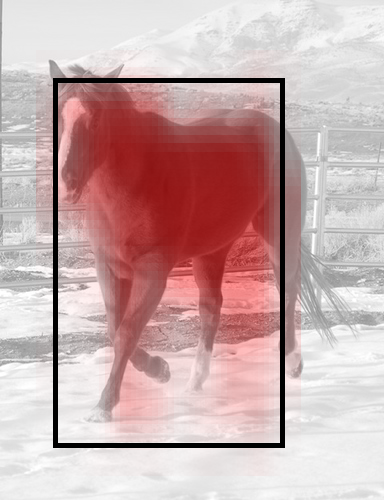}
\includegraphics[height=4\baselineskip]{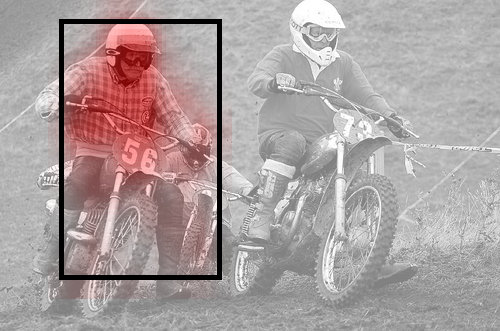}\\
\includegraphics[height=4\baselineskip]{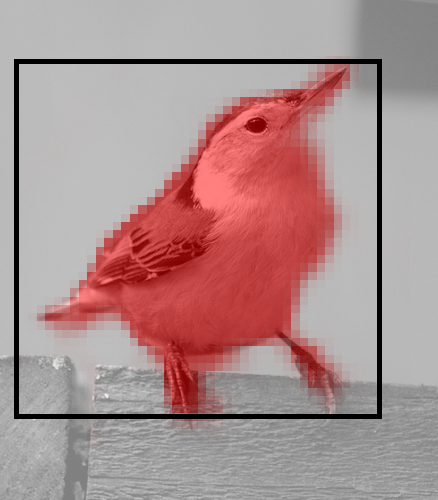}
\includegraphics[height=4\baselineskip]{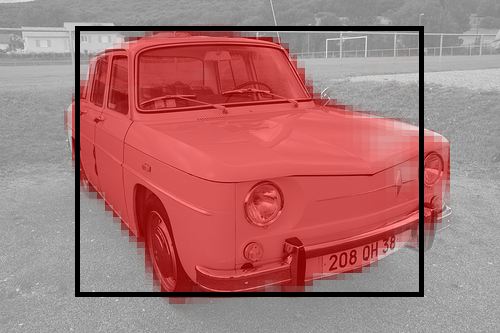}
\includegraphics[height=4\baselineskip]{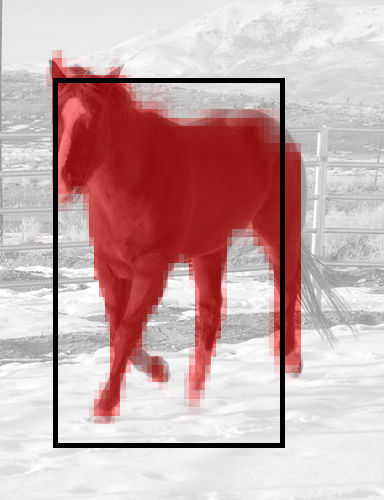}
\includegraphics[height=4\baselineskip]{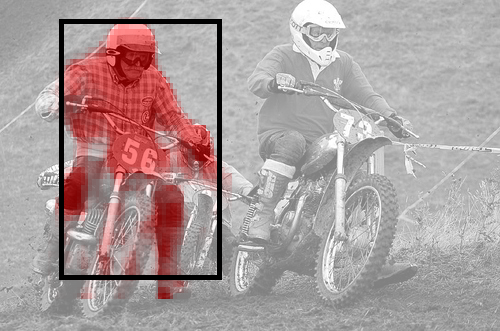}
\caption{Figure ground segmentations starting from bounding box detections. Top row: baseline using \emph{fc7}, bottom row: Ours.}\label{fig:SDS}
\end{figure}

To see how much of this performance is coming from the hypercolumn representation, we also run a baseline using just \emph{fc7} features. As expected, this baseline is only able to output a fuzzy segmentation, compared to the sharp delineation we get using hypercolumns. It performs considerably worse, losing 5 points at 0.5 overlap and almost 13 points at 0.7 overlap. Figure~\ref{fig:SDS} shows example segmentations.

We now replace the T-Net architecture for the O-Net architecture. This architecture is significantly larger, but provides an 8 point gain in detection AP~\cite{GirshickCVPR2014Arxiv}. We again retrain the R-CNN system using this architecture on MCG bounding box proposals. Again, for the hypercolumn representation we use the same network as the detector. We use the layers \emph{fc7}, \emph{conv4} with a neighborhood of 1 and \emph{pool3} with a neighborhood of 3. (We use \emph{pool3} instead of \emph{pool2} because the \emph{pool3} feature map has about half the resolution and is thus easier to work with.) 

We observe that the O-Net architecture is significantly better than the T-Net: we get a boost of 7.5 points at the 0.5 overlap threshold and 8 points at the 0.7 threshold. We also find that this architecture gives us the best performance on the SDS task so far: with simple bounding box detection followed by our hypercolumn-based mask prediction, we achieve a mean AP$^r$ of 56.5 at an overlap threshold of 0.5 and a mean AP$^r$ of 37.0 at an overlap threshold of 0.7. These numbers are about 6.8 and 11.7 points better than the results of~\cite{BharathECCV2014}. Last but not the least, we observe that the large gap between our hypercolumn system and the only-\emph{fc7} baseline persists, and is equally large for the O-Net architecture. This implies that the gain provided by hypercolumns is not specific to a particular network architecture. Figure~\ref{fig:SDS} visualizes our O-Net results.

We now implement the full pipeline proposed in Figure~\ref{fig:rescore}. For this, we expand the initial pool of detections as follows. We pick boxes with score higher than a threshold that were suppressed by NMS but that overlap the detections by less than 0.7. We then do a non-max suppression with a lenient threshold of 0.7 to get a pool of candidates to rescore. Starting from 20K initial detections per category across the dataset, our expanded pool is typically less than 50K per category, and less than 600K in total.

Next we segment each candidate using hypercolumns and score it using a CNN trained to classify regions. This network has the same architecture as O-Net. However, instead of a bounding box, this network takes as input the bounding box with the region background masked out. This network is trained as described in~\cite{BharathECCV2014}. We use features from the topmost layer of this network and concatenate them with the features from the top layer of the detection network, and feed these into an SVM. For training data, we use our expanded pool of candidates on the training set, and take all candidates for which segmentation predictions overlap groundtruth by more than 70\% as positive and those with overlap less than 50\% as negative. After rescoring, we do a non-max suppression using region overlap to get the final set of detections (we use an overlap threshold of 0.3). 

We get \textbf{60.0} mean AP$^r$ at 0.5, and \textbf{40.4} mean AP$^r$ at 0.7. These numbers are state-of-the-art on the SDS benchmark (in contemporary work, \cite{DaiCVPR2015} get slightly higher performance at 0.5 but do not report the performance at 0.7; our gains are orthogonal to theirs). Finally, on the semantic segmentation benchmark, we get a mean IU of \textbf{62.6}, which is comparable to state-of-the-art.
\begin{table*}
\begin{center}
\renewcommand{\arraystretch}{1.2}
\renewcommand{\tabcolsep}{1.0mm}
%\resizebox{\linewidth}{!}{
\footnotesize{
\begin{tabular}{lcccccccccccc}
\toprule
Metric & \cite{BharathECCV2014} & \cite{BharathECCV2014}  & Hyp & Hyp &Hyp+FT & Only & \emph{fc7}+ & \emph{fc7}+ & \emph{pool2}+ & $1\times 1$ & $2\times 2$ & $5 \times 5$\\
& &refined  & & +bbox-reg &+bbox-reg& \emph{fc7} & \emph{pool2} & \emph{conv4} & \emph{conv4} & grid & grid & grid\\
\midrule
mean AP$^r$ at 0.5 & 47.7 & 49.7  & 51.2 & 51.9 & \textbf{52.8} & 49.7 & 50.5 & 51.0 & 50.7 & 50.3 & 51.2 & 51.3\\
mean AP$^r$ at 0.7 & 22.8 & 25.3 & 31.6 & 32.4 & \textbf{33.7} & 25.8 & 30.6 & 31.2 & 30.8 & 28.8 & 30.2 & 31.8\\
\bottomrule
\end{tabular}
}
\end{center}
%}
\caption{Results on SDS on VOC2012 val using System 1. Our system (Hyp+FT+bbox-reg) is significantly better than~\cite{BharathECCV2014} (Section~\ref{sec:SDS1}).}\label{table:hypercolumn-basic}
\end{table*}

\begin{table}
\begin{center}
\footnotesize{
\renewcommand{\arraystretch}{1.2}
\begin{tabular}{lccccc}
\toprule
Metric & T-Net & T-Net & O-Net & O-Net & O-Net\\
&  Only & Hyp & Only & Hyp & Hyp+\\
&  \emph{fc7} & & \emph{fc7} & & Rescore\\
\midrule
mAP$^r$ at 0.5 & 44.0 & 49.1 & 52.6 & 56.5 & \textbf{60.0}\\
mAP$^r$ at 0.7 & 16.3 & 29.1 & 22.4 & 37.0 & \textbf{40.4}\\
\bottomrule
\end{tabular}
}
\end{center}
\caption{Results on SDS on VOC 2012 val using System 2. Our final pipeline is state-of-the-art on SDS. (Section~\ref{sec:SDS2})}\label{table:bbox}
\vspace{-3mm}
\end{table}

\section{Experiments on part localization}
We evaluate part localization in the unconstrained detection setting, where the task is to both detect the object and label its keypoints/segment its parts. This is different from most prior work on these problems~\cite{ToshevCVPR2014, TompsonNIPS2014, YamaguchiCVPR2012, YamaguchiICCV2013, BoCVPR2011, LuoICCV2013}, which operates on the immediate vicinity of ground-truth instances. We start from the detections of~\cite{BharathECCV2014}. We use the same features and network as in Section~\ref{sec:SDS1}. As before, we do all training on VOC2012 Train.
\paragraph{Keypoint prediction}\label{sec:keypoint}
\begin{table*}
\begin{center}
\footnotesize{
\renewcommand{\arraystretch}{1.2}
\begin{tabular}{lcccccccccccccc}
\toprule
Method & L.S & L.E & L.W & R.S & R.E & R.W & L.H & L.K & L.A & R.H & R.K & R.A & N & \textbf{Mean}\\
\midrule
\cite{GeorgiaBharathCVPR2014b} & 27.3 & 11.8 & 2.8 & 27.1 & 12.2 & 3.4 & \textbf{11.4} & 4.9 & 3.2 & \textbf{10.6} & 4.4 & 3.8 & 42.9 & \emph{12.8}\\
\cite{GkioxariArxiv2014} & 32.1 & 14.6 & 5.6 & 32.5 & \textbf{16.6} & 5.9 & 10.8 & 4.8 & 4.8 & 9.7 & 4.0 & 4.6 & 52.0 & \emph{15.2}\\
Only \emph{fc7} & 22.7 & 9.9 & 2.8 & 25.5 & 10.0 & 2.6 & 6.6 & 3.5 & 5.2 & 7.7 & 3.4 & 4.2 & 34.0 & \emph{10.6}\\
Hyp & 32.2 & 16.5 & 11.5 & 31.2 & 16.6 & 9.3 & 9.6 & 7.1 & \textbf{9.1} & 8.0 & 4.2 & \textbf{8.2} & 57.5 &\emph{17.0}\\
Hyp+FT & \textbf{33.7} & \textbf{21.9} &\textbf{12.3} & \textbf{35.2} & \textbf{20.9} & \textbf{15.3} & 6.7 &\textbf{7.7} &8.1 & \textbf{9.1} &\textbf{5.6} & 6.1 & \textbf{58.4} & \emph{\textbf{18.5}}\\
\bottomrule
\end{tabular}
}
\end{center}
\caption{Results on keypoint prediction (APK on the Person subset of VOC2009 val). Our system is 3.3 points better than~\cite{GkioxariArxiv2014} (Section~\ref{sec:keypoint}).}\label{table:apk}
\end{table*}
We evaluate keypoint prediction on the ``person" category using the protocol described in~\cite{GeorgiaBharathCVPR2014b}. The test set for evaluating keypoints is the person images in the second half of VOC2009 val. We use the APK metric~\cite{YangTPAMI2013}, which evaluates keypoint predictions in a detection setting. Each detection comes with a keypoint prediction and a score. A predicted keypoint within a threshold distance (0.2 of the torso height) of the ground-truth keypoint is a true positive, and is a false positive otherwise. The area under the PR curve gives the APK for that keypoint.

We start from the person detections of~\cite{BharathECCV2014}. We use bounding box regression to start from a better bounding box. As described in Section~\ref{sec:hypercolumn} we train a separate system for each keypoint using the hypercolumn representation. We use keypoint annotations collected by~\cite{BourdevECCV2010}. We produce a heatmap for each keypoint and then take the highest scoring location of the heatmap as the keypoint prediction. 

The APK metric requires us to attach a score with each keypoint prediction. This score must combine the confidence in the person detection and the confidence in the keypoint prediction, since predicting a keypoint when the keypoint is invisible counts as a false positive. For this score we multiply the value of the keypoint heatmap at the predicted location with the score output by the person detector (which we pass through a sigmoid).

\begin{figure}
\centering
\includegraphics[height=4\baselineskip]{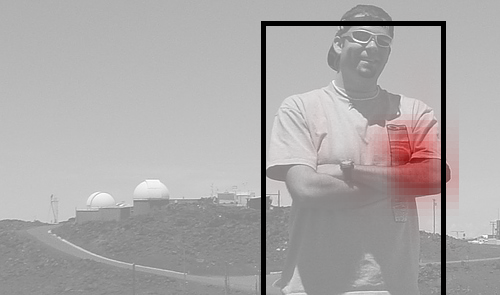}
\includegraphics[height=4\baselineskip]{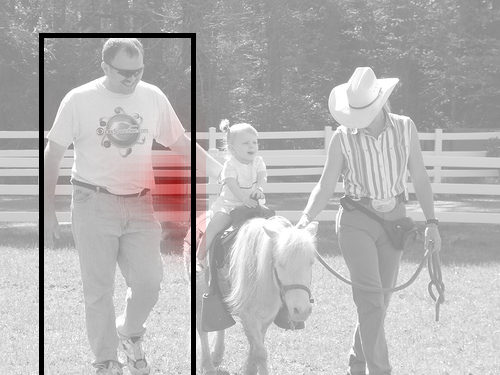}
\includegraphics[height=4\baselineskip]{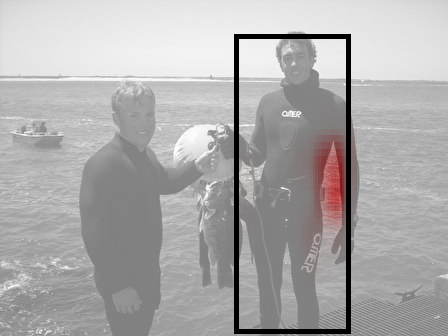}\\
\includegraphics[height=4\baselineskip]{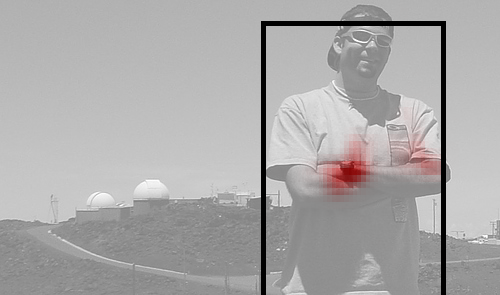}
\includegraphics[height=4\baselineskip]{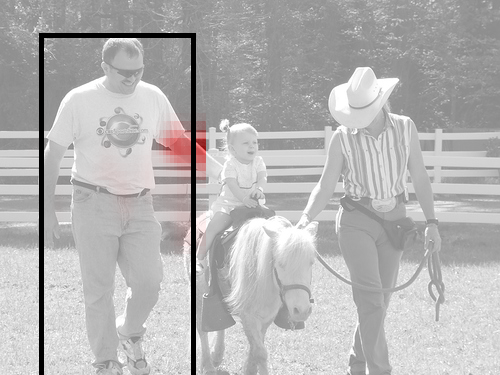}
\includegraphics[height=4\baselineskip]{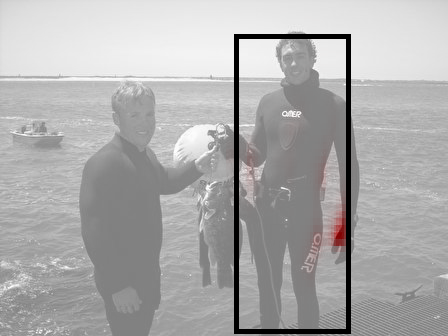}
\caption{Keypoint prediction (left wrist). Top row: baseline using \emph{fc7},  bottom row: ours (hypercolumns without finetuning). In black is the bounding box and the predicted heatmap is in red. We normalize each heatmap so that the maximum value is 1.}\label{fig:kp}
\end{figure}

Results are shown in Table~\ref{table:apk}. We compare our performance to~\cite{GkioxariArxiv2014}, the previous best on this dataset. Gkioxari et al.~\cite{GkioxariArxiv2014} finetuned a network for pose, person detection and action classification, and then trained an SVM to assign a score to the keypoint predictions. Without any finetuning for pose, our system achieves a 1.8 point boost. A baseline system trained using our pipeline but with just the \emph{fc7} features performs significantly worse than our system, and is even worse than a HOG-based method \cite{GeorgiaBharathCVPR2014b}. This confirms that the gains we get are from the hypercolumn representation. Figure~\ref{fig:kp} shows some example predictions. 

Finetuning the network as described in Section~\ref{sec:hypercolumn} gives an additional \textbf{1.5} point gain, raising mean APK to \textbf{18.5}. \vspace{-1mm}%The improvements are especially large for keypoints on the arm.
%
%
%Note that there are more powerful keypoint prediction algorithms proposed in the literature, but these typically show results on a tightly cropped ground truth bounding box, whereas we are interested in the setting where one has to both detect the person and identify their keypoints. 

\vspace{-0.3cm}
\paragraph{Part labeling}\label{sec:partlabel}
\begin{table}
\begin{center}
%\footnotesize{
\resizebox{\linewidth}{!}{
\renewcommand{\arraystretch}{1.2}
\begin{tabular}{lccccccc}
\toprule
AP$^r_{part}$ at 0.5 & Person & Horse & Cow & Sheep & Cat & Dog & Bird\\
\midrule
Only \emph{fc7} & 21.9 & 16.6 & 14.5 & 38.9 & 19.2 & 8.5 & \textbf{15.4}\\
Hyp & \textbf{28.5} & \textbf{27.8} & \textbf{21.5} & \textbf{44.9} & \textbf{30.3} & \textbf{14.2} & 14.2\\
\bottomrule
\end{tabular}
%}
}
\end{center}
\caption{Results on part labeling. Our approach (Hyp) is almost uniformly better than using top level features (Section~\ref{sec:partlabel}).}\label{table:parsing}
\vspace{-3mm}
\end{table}
We evaluate part labeling on the articulated object categories in PASCAL VOC: person, horse, cow, sheep, cat, dog, bird. We use the part annotations provided by~\cite{ChenCVPR2014}. We group the parts into  top-level parts: head, torso, arms and legs for person, head, torso, legs and tail for the four-legged animals and head, torso, legs, wings, tail for the bird. We train separate classifiers for each part. At test time, we use the Hyp+bbox-reg+FT system from Section~\ref{sec:SDS1} to predict a figure-ground mask for each detection, and to every pixel in the figure-ground mask, we assign the part with the highest score at that pixel. 

\begin{figure}
\centering
\includegraphics[height=4\baselineskip]{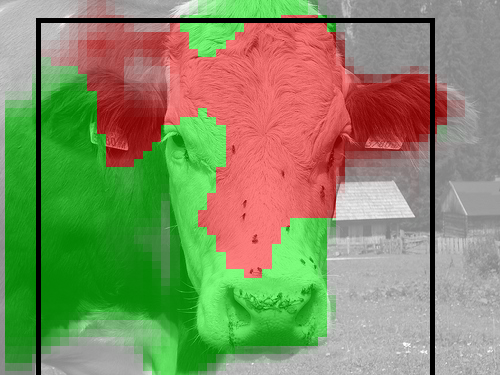}
\includegraphics[height=4\baselineskip]{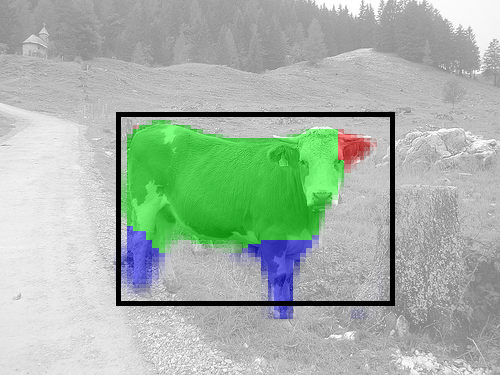}
\includegraphics[height=4\baselineskip]{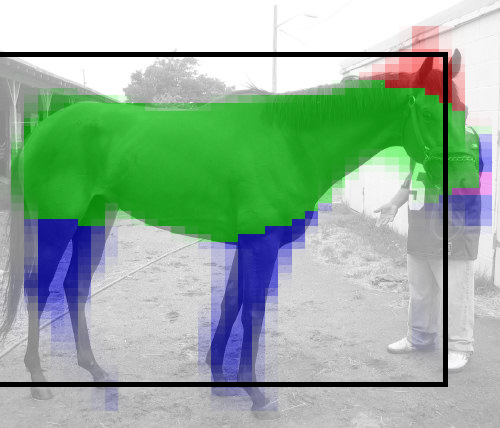}
\includegraphics[height=4\baselineskip]{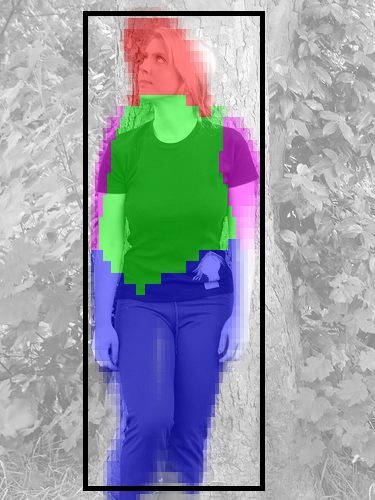}\\
\includegraphics[height=4\baselineskip]{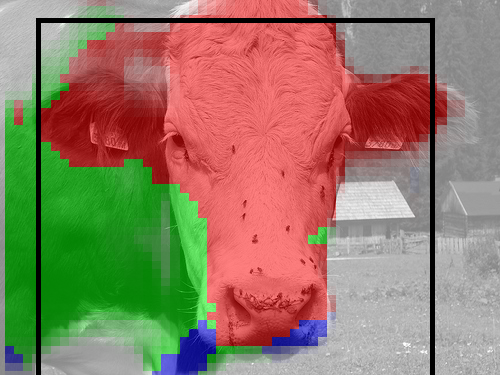}
\includegraphics[height=4\baselineskip]{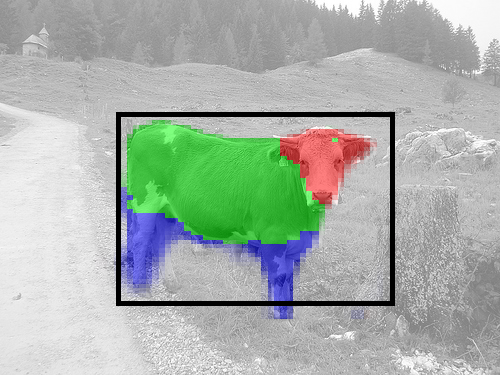}
\includegraphics[height=4\baselineskip]{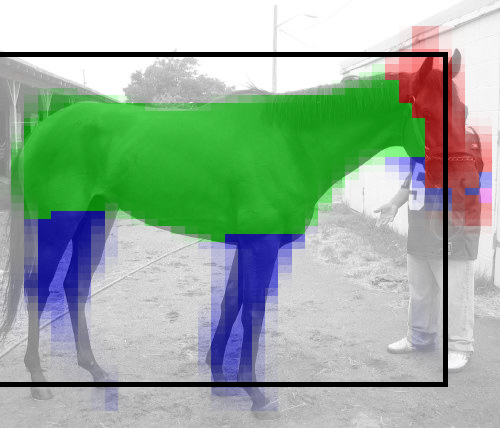}
\includegraphics[height=4\baselineskip]{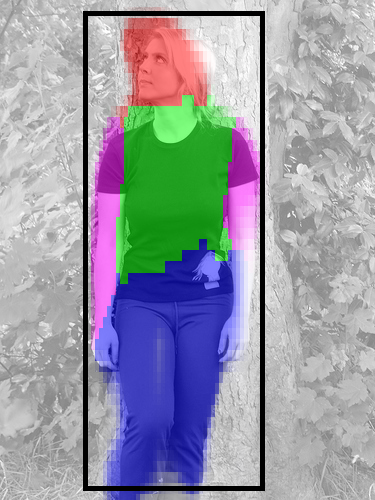}
\caption{Part labeling. Top: baseline using \emph{fc7}, bottom: ours (hypercolumns). Both rows use the same figure-ground segmentation. Red: head, green: torso, blue: legs, magenta: arms.}\label{fig:part}
\vspace{-0.5cm}
\end{figure}

For evaluation, we modify the  definition of intersection-over-union in the AP$^r$ metric~\cite{BharathECCV2014}: we count in the intersection only those pixels for which we also get the part label correct. We call this metric AP$^r_{part}$. As before, we evaluate both our system and a baseline that uses only \emph{fc7} features. Table~\ref{table:parsing} shows our results. We get a large gain in almost all categories by using hypercolumns. Note that this gain is entirely due to improvements in the part labeling, since both methods use the same figure-ground mask. Figure~\ref{fig:part} shows some example part labelings.
\vspace{-2.5mm}

%%%%Conclusion
\nopagebreak
\section{Conclusion}
We have shown that the hypercolumn representation provides large gains in three different tasks. We also believe that this representation might prove useful for other fine-grained tasks  such as attribute or action classification. We leave an investigation of this to future work.

\clearpage
\noindent \textbf{Acknowledgments}. This work was supported by ONR MURI N000141010933, a Google Research
Grant and a Microsoft Research fellowship. We thank NVIDIA for providing GPUs
through their academic program.

{\small
\bibliographystyle{ieee}
\bibliography{egbib}
}
\end{document}